\documentclass[10pt,twocolumn,letterpaper]{article}
\usepackage{iccv}
\usepackage{times}
\usepackage{epsfig}
\usepackage{graphicx}
\usepackage{amsmath}
\usepackage{amssymb}
\usepackage{booktabs}
\usepackage{algpseudocode}
\usepackage{chngcntr}
\usepackage{colortbl} 
\usepackage[table,xcdraw]{xcolor}
\usepackage{tabularx, threeparttable}
\usepackage{multirow}

\usepackage[accsupp]{axessibility}

\usepackage[ruled,linesnumbered]{algorithm2e}
\usepackage[accsupp]{axessibility}

\usepackage[pagebackref=true,breaklinks=true,letterpaper=true,colorlinks,bookmarks=false]{hyperref}

\iccvfinalcopy
\def\iccvPaperID{2376} % *** Enter the ICCV Paper ID here

\begin{document}
%%%%%%%%% TITLE
\title{Exploring Model Transferability through the Lens of Potential Energy}
\author{
{Xiaotong Li{$^{1,2}$}} \qquad Zixuan Hu{$^{1}$} \qquad Yixiao Ge{$^{3}$} \qquad Ying Shan{$^{3}$} \qquad Ling-Yu Duan{$^{1,2}$}\thanks{Corresponding Author.} \\
% \institute{
\normalsize
$^{1}$	School of Computer Science, Peking University, Beijing, China, \\
\normalsize 
$^{2}$ Peng Cheng Laboratory, Shenzhen, China,~~ $^{3}$ ARC Lab, Tencent PCG, Beijing, China \\
{\tt\small{lixiaotong@stu.pku.edu.cn, \{yixiaoge, yingsshan\}@tencent.com}},\\
\tt\small{\{hzxuan, lingyu\}@pku.edu.cn}
% }
}
% \author{%
% \textbf{Manyi Zhang$^{1,}$\thanks{Equal contributions. This work was completed when the first two authors were interns guided by the last author.}
% \quad 
% Xuyang Zhao$^{2,*}$
% \quad
% Jun Yao$^{3}$
% \quad Chun Yuan$^{1,}$\thanks{Correspondance to Weiran Huang (weiran.huang@outlook.com) and Chun Yuan (yuanc@sz.tsinghua.edu.cn).}
% \quad
% Weiran Huang$^{4,\dag}$} \\[0.3cm] 
% $^{1}$SIGS, Tsinghua University \quad $^{2}$Peking University \quad 
% $^{3}$Huawei Noah's Ark Lab \\[0.1cm] 
% $^{4}$Qing Yuan Research Institute, SEIEE, Shanghai Jiao Tong University
% }
\maketitle
\ificcvfinal\thispagestyle{empty}\fi
\begin{abstract}

% In computer vision, transfer learning has become a critical milestone due to the proliferation of pre-trained deep learning models. 
Transfer learning has become crucial in computer vision tasks due to the vast availability of pre-trained deep learning models. However, selecting the optimal pre-trained model from a diverse pool for a specific downstream task remains a challenge. Existing methods for measuring the transferability of pre-trained models rely on statistical correlations between encoded static features and task labels, but they overlook the impact of underlying representation dynamics during fine-tuning, leading to unreliable results, especially for self-supervised models. In this paper, we present an insightful physics-inspired approach named PED to address these challenges. We reframe the challenge of model selection through the lens of potential energy and directly model the interaction forces that influence fine-tuning dynamics. By capturing the motion of dynamic representations to decline the potential energy within a force-driven physical model, we can acquire an enhanced and more stable observation for estimating transferability. The experimental results on 10 downstream tasks and 12 self-supervised models demonstrate that our approach can seamlessly integrate into existing ranking techniques and enhance their performances, revealing its effectiveness for the model selection task and its potential for understanding the mechanism in transfer learning. Code will be available at \href{https://github.com/lixiaotong97/PED}{https://github.com/lixiaotong97/PED}.

\end{abstract}

%%%%%%%%% BODY TEXT
\section{Introduction}
\counterwithout{figure}{section}

Transfer learning has achieved remarkable success in computer vision by fine-tuning models pre-trained on large-scale datasets (\textit{e.g.,} ImageNet \cite{imagenet}) for downstream tasks.
% leveraging models pre-trained on large-scale datasets (\textit{e.g.,} ImageNet \cite{imagenet}) and fine-tuning them for the downstream tasks. 
However, the proliferation of various network designs and training strategies presents a challenge in selecting an optimal model from the extensive range of options for a particular downstream task.
While fine-tuning each potential model in a brute-force manner is a direct approach for model selection, it is computationally infeasible due to the growing number of model candidates.
\begin{figure}[t]
\begin{center}
\includegraphics[width=1.0\linewidth]{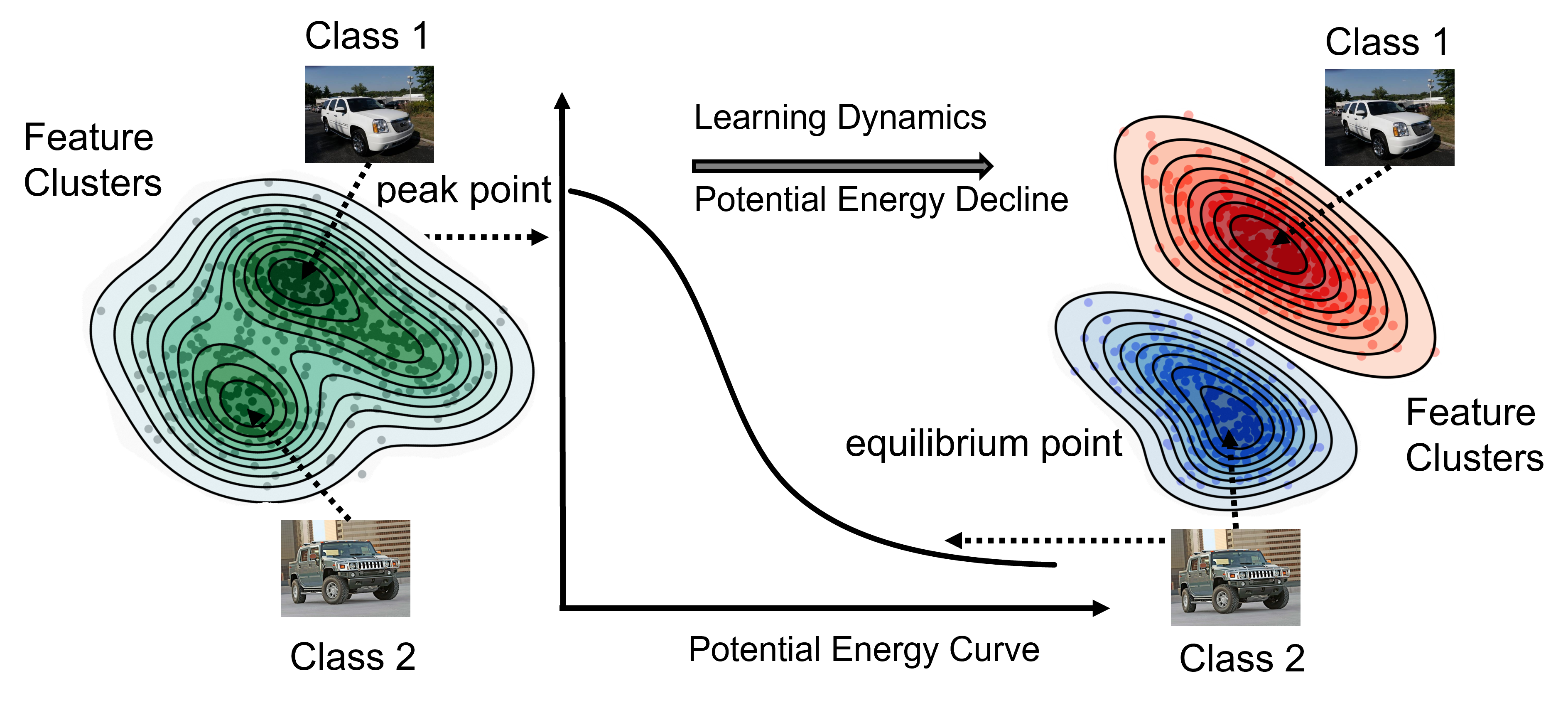}
%\framebox[4.0in]{$\;$}
%\fbox{\rule[-.5cm]{0cm}{4cm} \rule[-.5cm]{4cm}{0cm}}
\end{center}
\caption{We analogy the physical concept and consider the transfer learning dynamics in the perspective of potential energy. The objective to push apart different classes can be viewed as an interaction ``force'' to decline the system ``potential energy'' and the dynamics can be seen as a process from unstable to stable point of the energy plane.
}
% Once the label information gets involved in the feature space of target dataset, we can model there is some kind of physical force to push away the samples of different class. .}
\label{fig:motivation}
\vspace{-5pt}
\end{figure}

To address this challenge, prior studies~\cite{nleep,logme,sfda,gbc} have endeavored to efficiently measure the transferability of pre-trained models related to the separability of encoded representations. The principle underlying these approaches is to select a pre-trained model that can effectively segregate its initial features using the ground-truth labels (\textit{i.e.}, image classes) in the downstream task.

While the aforementioned methodology is effective for ranking supervised pre-trained models, which are originally optimized toward class separability, it is not always reliable for ranking un/self-supervised pre-trained models~\cite{ericsson2021well}. These models have emerged as dominant in transfer learning and have exhibited superior performance compared to supervised learning models. Nevertheless, self-supervised models exhibit different properties due to the discrepancy between pre-training target and downstream classification objective \cite{ericsson2021well}. We argue that the limitations of the existing separability-based methodology stem from its inability to consider the underlying representation dynamics during the fine-tuning process of transfer learning and encounter challenges for ranking self-supervised models. 

\begin{figure*}[t]
\begin{center}
\label{fig:pipeline}
\includegraphics[width=1.0\textwidth]{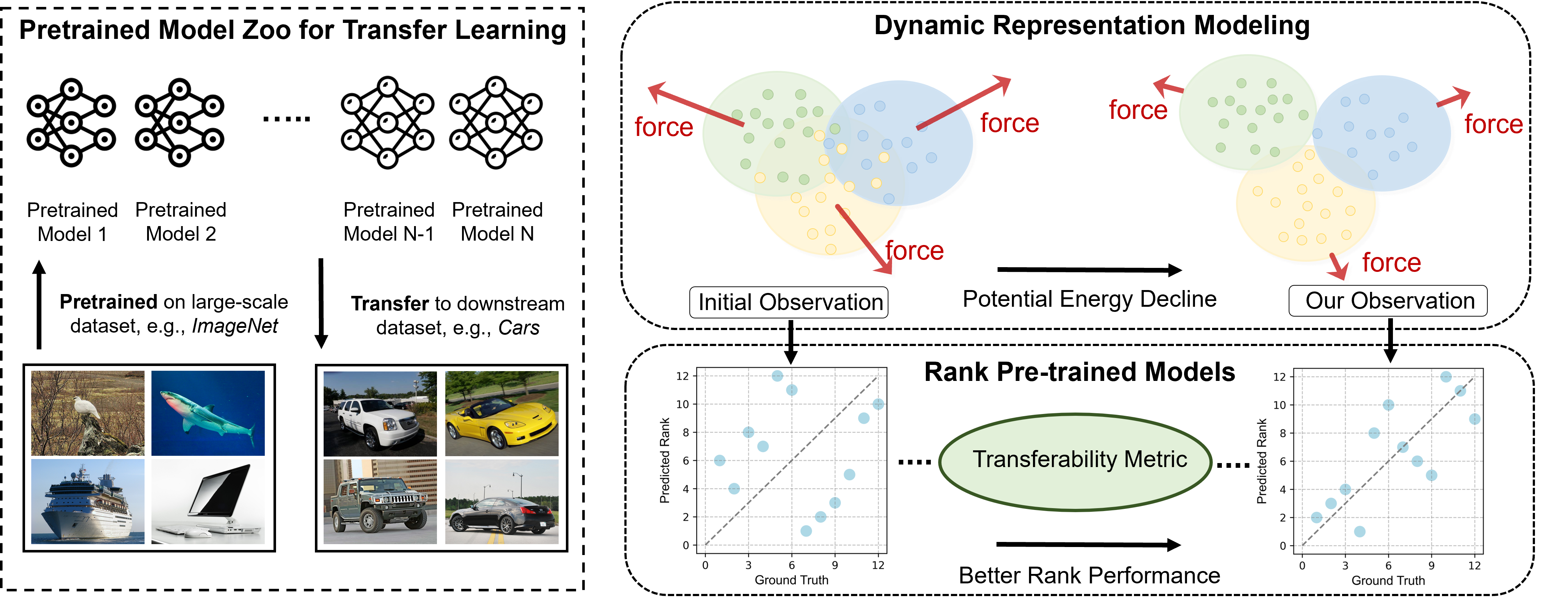}
%\framebox[4.0in]{$\;$}
%\fbox{\rule[-.5cm]{0cm}{4cm} \rule[-.5cm]{4cm}{0cm}}
\caption{Pipeline of the proposed Potential Energy Decline (PED) approach for model selection. The models are first trained on a large-scale dataset and then transferred to a given downstream task. We propose a novel approach through the perspective of potential energy to alleviate the limitations of the initial observations. By treating the learning dynamics as minimizing the potential energy and considering the system's tendency to change, we model the interaction force of different clusters using a repulsion-based force to capture the moving tendency. Subsequently, we unfreeze the state of the start point and apply the force to push each class away, leading to a decline of the potential energy. This approach leads to a more stable observation of features, resulting in more accurate transferability score predictions.}
\end{center}
\vspace{-0.5cm}
\end{figure*}
Modeling the representation dynamics for model ranking is a crucial yet challenging task. The present study focuses on image classification tasks without loss of generality.
% Without loss of generality, the paper considers the typical image classification tasks.
% Let us first rethink the nature of model optimization during transfer learning, where backward-propagating the gradients measured by classification cross-entropy loss seeks to cluster features out of the same classes. 
To understand the nature of model evolution in transfer learning, we examine the process of backward-propagating gradients measured by classification cross-entropy loss. The process aims to cluster features out of the same classes, which can be viewed as creating a force that separates the clusters and system potential energy gets decreased driven by the force from a physical perspective~\cite{uniformface,regularface}. 
Reframing model evolution through the lens of potential energy reveals that the pre-trained model attains a state of equilibrium after pre-training, with low interaction forces and stable sample relationships. However, this stable state is disrupted when the model is transferred to a downstream task, leading to changes in the potential energy plane. Intuitively, predicting model transferability based on an unstable observation will hinder its predicting performance. Drawing from the principles of physics \cite{energyprinciple}, the present unstable state is inclined to move towards a reduction in potential energy and results in a more stable state. To properly predict a model's transferability, it is essential to model the force that determines the system's tendency.

%To properly predict a model's transferability, it is essential to model the force that determines the system's tendency.
%to change from one state to more stable state, drawing from the principles of physics.
% We seek to quantify the force acting on each class cluster and measures its corresponding movement on the potential energy landscape implicitly defined by the loss function.
% Specifically, we model the loss function as implicitly defining a potential energy plane, with the interaction between classes represented as a force that reduces the potential energy of the system on the latent space. 
% Specifically, we consider each class's features in the downstream task as a ball in the latent space, with the class center indicating the coordinate and the variation representing the radius. The interaction force between different classes can be formulated by the overlap radius of the two balls. 
% By observing the moving tendency after unfreezing the system, we can simulate the positions of features and obtain a new system state with lower potential energy. 
% By unfreezing the system and observing the moving tendency, we can readily simulate the positions of features without backward-propagation, rendering a new stable state with lower potential energy. 
% By integrating our force-directed dynamic representations with an off-the-shelf ranking algorithm, \textit{e.g.}, LogME \cite{logme}, we obtain a transferability score for each pre-trained model that reflects its potential for use in a given downstream task. 

We therefore formulate the representation dynamics in terms of potential energy\footnote{Throughout this paper, the ``energy'' is a quantitative property held by features because of relative class positions in its latent space.} and propose the approach to tackle these challenges named Potential Energy Decline (PED), as demonstrated in Fig.\ref{fig:pipeline}. To quantify the interaction force acting on each class cluster and measure its corresponding movement on the potential energy landscape implicitly defined by the optimization objective, we consider each class's representations in the downstream task as a ball in the latent space, with the class center indicating the coordinate and the variation representing the radius. The interaction force between different classes is formulated by the overlap radius of the two balls. We can simulate the positions of dynamic representations without backward-propagation by unfreezing the system and observing the moving tendency that leads to a new state with lower potential energy. Our force-directed dynamic representations provide a better observation and can be readily integrated into existing ranking algorithms, such as LogME \cite{logme}, to achieve better model transferability measurement.

To the best of our knowledge, we are the first to explore model transferabilty through the lens of potential energy and simulate the underlying representation dynamics during transfer learning in a physics-driven approach.
To evaluate our proposed method, we conduct extensive experiments on a variety of self-supervised pre-trained models. Our method can be easily integrated into existing approaches with negligible time consumption. The experimental results on 10 downstream tasks and 12 self-supervised pre-trained models demonstrate our method can boost various metrics for more accurate prediction.
Our findings might have implications beyond the realm of image classification, as our approach is generic and can be extended to more pre-trained models and other downstream tasks. We hope that our work will inspire future studies and have a broader impact in the field of transfer learning.
% Our experimental results demonstrate that our method consistently improves the accuracy of model ranking, and our method can be easily integrated into existing approaches with negligible time consumption. 
% Finally, we hope that our work sheds light on the importance of considering transfer learning dynamics from the perspective of potential energy.

\section{Related work}

% \subsection{Transfer Learning}
% In the field of computer vision, transfer learning has become a significant milestone due to the availability of a model zoo of pre-trained deep learning models. As the number of pre-training strategies continues to expand, the model zoo is becoming increasingly extensive, with numerous pre-trained models available for use, such as Pytorch Model Zoo and HuggingFace \cite{timm}. As a result, selecting the most appropriate model from the model zoo for a particular downstream task has become a significant challenge. The goal of this paper is to address this issue by proposing a method for selecting the optimal model from the model zoo that is well-suited to the given task.

\subsection{Transferability Metric}
In the field of computer vision, transfer learning has become a significant milestone due to the availability of a model zoo of pre-trained deep learning models \cite{ericsson2021well,timm}. As a result, selecting the most appropriate model from the model zoo for a particular downstream task has become an important challenge and model selection is therefore proposed to tackle this problem with a low budget estimation.
%it has received much attention from researchers in recent years \cite{leep,NCE,logme,nleep,sfda,h-score,gbc,parc,rsa,task2vec,taskonomy}. 

%To address the problem of model selection, a variety of methods have been proposed, such as LEEP \cite{leep}, NCE \cite{NCE}, LogME \cite{logme}, $\mathcal{N}$LEEP \cite{nleep}, SFDA \cite{sfda}, H-Score \cite{h-score}, GBC \cite{gbc}, TransRate, PARC \cite{parc}, RSA \cite{rsa}, Task2Vec \cite{task2vec}, and Taskonomy \cite{taskonomy}. 
%\paragraph{Model Transferability. }
\paragraph{Transferabilty Metric.} Model transferability is a fundamental aspect in the field of transfer learning, and it has received much attention from researchers in recent years for designing various transferability metrics \cite{leep,NCE,logme,nleep,sfda,h-score,gbc,parc,rsa,task2vec,taskonomy,pactran,agostinelli2022stable,ctc}. For example, LEEP \cite{leep} estimates the joint probability of the source and target label space, while NLEEP \cite{nleep} predicts the label by fitting a Mixture of Gaussian model. LogMe \cite{logme} propose to estimate the maximum value of label evidence given the encoded features. PARC \cite{parc} uses pairwise pearson product-moment correlation between the features of
each pair of images. GBC \cite{gbc} measures the pairwise class overlaps in distribution density with a Bhattacharyya coefficient. SFDA \cite{sfda} measures model transferability using the class discrimination in a Fisher space and proposes a self-challenging approach named ConfMix to simulate the hard negative samples in fine-tuning. These methods have made significant contributions to the field of transfer learning, but there is still challenge to deal with un/self-supervised pre-trained models \cite{byol,swav,simclr-v1,simclr-v2,pcl,deepcluster}, because the models be useful as a starting point for many downstream tasks, they are not sufficient on their own to separate different classes of samples \cite{ericsson2021well}.

\subsection{Energy-based Methods in Deep Learning}
%------------------------------------------------------------------------
% \mathcal{M}

Energy-based methods have a long-standing history in the field of machine learning and have been commonly employed to model interactions between objects \cite{rbm,DeepBM}. 
%By representing the image or feature space as a physical system with interacting particles, the potential energy of the system can be minimized to find the optimal solution. 
Early works in this area can be traced back to Restricted Boltzmann Machines (RBM) \cite{rbm} and DeepBM \cite{DeepBM}, that use a series of layers of stochastic binary units to represent data and is trained to minimize an energy function measured by the compatibility between the input data and the internal representation. The Hopfield network \cite{hopfield2007hopfield} is also designed to find a state of minimum energy, which corresponds to a stable solution or equilibrium and is used for segmentation \cite{rout1998multi}. In face recognition task, Uniformface\cite{uniformface} and Regularface \cite{regularface} also borrow the force in potential energy to model the inter-class regularization to design a optimized loss. Inspired by these works, we reframe the challenge of ranking self-supervised pre-trained model through the lens of potential energy. To the best of our knowledge, we are the first to consider transfer learning from an energy-based perspective and propose a physical approach to model the dynamic representation in model selection task.
%which has been applied to some vision tasks, such as object detection, semantic segmentation, and image retrieval \cite{uniformface, regularface}. This approach has shown promising results in improving the accuracy and robustness of deep learning models, as well as providing a new perspective for understanding the behavior of neural networks. The potential energy can be abstracted as a function of current system state. For example, the optimization of the deep learning networks can be viewed as minimizing the energy based model by stochastic gradient descent. Energy-based model and JFM \cite{Grathwohl2020Your} imply that the classification task has the relation with energy.
%However, directly optimizing the 

\section{Methodology}
In this section, we first present the problem setup, ranking metrics, and evaluation protocol of the model selection problem. Then we state the inspiration from a physical view and illustrate how we efficiently model the representation dynamics in terms of potential energy.
Without loss of generality, we take classification as an example throughout our paper.
% are motivated to propose the method. 

\subsection{Preliminaries}
\label{preliminaries}

\textbf{Problem Setup.} 
% Consider a pre-trained model $\Phi_{i}$ from a model zoo ${\{{\Phi_i}\}_{i=1}^{N}}$ that is transferred to the downstream dataset $\mathcal{T}=\{X,Y\}$. 
Consider a model zoo ${\{{\Phi_i}\}_{i=1}^{N}}$ from which the selected pre-trained model can be transferred to a downstream dataset $\mathcal{T}=\{X,Y\}$. 
The purpose of model selection is to predict model transferability with minor computational costs without fine-tuning.
% performance rank from the model zoo with a low budget cost, instead of fine-tuning the models.

\textbf{Ranking Metric.} Given a model $\Phi_{i}$, we encode the features $Z$ for the downstream dataset $X$, then feed the features and labels to a metric $\mathcal{M}(Z, Y)$ to estimate a transferability score $P_i$. Intuitively, the metric measures the transferability based on the separability of the encoded features.
% certain compatibility between labels and extracted features.

% $\{{P_i}\}_{i=1}^{N}$ of each model, where the model selection predict the model transferability based on certain compatibility between labels and extracted features.

\textbf{Evaluation Protocol.} To evaluate different model selection algorithms, we follow previous arts to estimate the weighted Kendalls' $\tau_w$ \cite{logme} between ground-truth model rankings and the predicted rankings. Specifically, we obtain the ground-truth rankings $\{{G_i}\}_{i=1}^{N}$ via fully fine-tuning. Then $\tau_w$ can be formulated as
% pre-calculate the performance ranking of each model by fully fine-tuning to get the ground truth transferability score $\{{G_i}\}_{i=1}^{N}$. Then, the well-known rank correlation measurement weighted Kendalls' $\tau_w$ \cite{logme} is adopted to evaluate the accuracy of model selection metric.
 \begin{equation}
     \tau_w=\frac{2}{N(N-1)}\sum_{1\leq i \textless j \leq N}\text{sign}(G_i-G_j)\cdot \text{sign}(P_i-P_j).
 \end{equation}

% (TO BE CONTINUED) Given the estimated model transferability scores, we can select the optimal pre-trained model for a particular downstream task. 
Although existing ranking methods \cite{logme,nleep,sfda,gbc,parc} are effective for supervised pre-trained models, they are not always reliable
for un/self-supervised pre-trained models which are not trained toward class separability and need to be fine-tuned for downstream tasks. 
We argue that it is actually due to the fact that they mostly ignore the underlying representation dynamics during the fine-tuning process of transfer learning. Properly and efficiently modeling such dynamics is called for.

% In the field of vision pre-training, self-supervised learning has gained prominence as a dominant approach in recent years, demonstrating superior transferability compared to supervised learning methods. Nevertheless, the potential learning dynamics can significantly impact the performance of traditional model transferability prediction metrics for self-supervised learning models. In contrast, supervised models exhibit relatively small and less prominent dynamic movement due to the high coincidence degree of the supervised classification objective with downstream tasks. Therefore, in this paper, we analyze the transferability prediction of self-supervised learning methods to evaluate our proposed approach.
% In contrast, due to the supervised classification objective has a high coincidence degree to downstream task, the system movement of supervised model is relative small and less prominent. 

\subsection{Understanding Transfer Learning from a Dual View}

It is of great significance to rethink the fine-tuning optimization process and dive into the dynamics of learning representations. In this section, we provide a novel dual view to reframe gradient-based optimization into a physics perspective.

% It is of great significance to rethink the model selection task and dive into the dynamics of learning representations. In this section, we provide a dual view to consider the representation dynamic in deep learning and physics perspectives.
% In transfer learning, a model is first pre-trained on a source dataset and then fine-tuned on a downstream labeled dataset. During pre-training, the model $\Phi(\theta)$ is optimized with a specific training objective and will be stored in the pre-trained model repository when the pre-training stage is complete. 

\subsubsection{A Gradient-based View}

When transferring a pre-trained model $\Phi$ parameterized by $\theta$ to a particular downstream task $\{X,Y\}$, the model is generally optimized toward minimizing a loss function $\mathcal{L}$ (\textit{e.g.}, cross-entropy loss)
% In the transfer stage for a downstream task, a labeled dataset $\{X,Y\}$ is used to minimize a new training objective $\mathcal{L}$ 
through gradient backpropagation, such as
%In physics, there exists a relationship between potential energy and its force, which meets:
\begin{equation}
\label{eq:optimization}
\begin{split}
    &\theta^{t+1}=\theta^t-\frac{\partial \mathcal{L}(Z^t,Y)}{\partial \theta^t}, \\
    &Z^{t}=\Phi(X|\theta^{t}).
\end{split}
\end{equation}
As a result of the iterative optimization in Eq. (\ref{eq:optimization}), the ability of $\Phi$ to discriminate between different classes has been improved. The separability of different classes in the latent space has also been enhanced.
% with updated network. 
The underlying representation dynamics can be thought of as the state evolution from $Z^0$ to $Z^T$, where $T$ is the total number of iterations. Obviously, optimizing the network and updating its encoded features is non-trivial for the model selection process, and what we need is to simulate the dynamics representation without resorting to fine-tuning.
% The previous paragraph describes the learning representation dynamics from an optimization perspective. However, updating the model to encode features is not a trivial task in the model selection process. 

\subsubsection{An Energy-based View}
% Taking cues from prior works \cite{rbm, DeepBM, regularface, uniformface}, 
Every coin has two sides, we discover that the learning objective in optimization shows resemblances to the concept of potential energy in physics. Intuitively, the loss function $\mathcal{L}$ and the gradient $\frac{\partial \mathcal{L}}{\partial \theta}$ show similarities in form with potential energy $U$ and force $F$, \textit{i.e.}, $-\frac{\partial U}{\partial s}$, respectively. 
% The loss gradient acts on the network parameters to minimize the loss and distinguish between different class features, similar to how the gradient of potential energy with respect to position determines the force acting on an object to decrease potential energy.
% The gradients work to minimize the loss and distinguish between different class features by adjusting the network parameters. This process is analogous to the way the force acting on an object is determined by the gradient of potential energy with respect to position, causing the object to move towards lower potential energy.
The loss gradient minimizes the loss and distinguishes between different class features by adjusting the network parameters, in a similar way to how an object's position affects the force acting on it to decrease potential energy. Building upon this insight, we reformulate the optimization process during transfer learning from the lens of potential energy in physics.

% In physics, the movement of objects under the influence of an interaction force $F$ can be described by the direction $n$, along which the samples move, and the potential energy of the system decreases accordingly, as shown in the equation:
From the physics perspective, the direction of object movement under the influence of an interaction force $F$ can be denoted by the path $n$. As the object moves in the path of $n$, the potential energy of the system decreases, as expressed in the following equation:
\begin{equation}
\begin{split}
&U(Z^{t+1})=U(Z^{t})-\int_n Fds, \\
&Z^{t+1}=Z^{t}+\int_n ds,
\end{split}
\end{equation}
where 
% $F$ represents the force to decline the potential energy and 
$\int_n ds$ is the relative position movement along the path of $n$. By viewing optimization in terms of physics, we shed light on the behavior of the loss function and the optimization process. The concepts of potential energy and network gradients can be seen as two sides of the same coin, which can help us understand the nature of the optimization process and model the representation dynamics without loss backpropagation.
\paragraph{Rethinking transfer learning from the energy view.}
Based on the above findings, we propose to revisit transfer learning through the lens of potential energy. When the pre-trained model converges, the model ``system'' defined by the training objective reaches a state of relative stability with equilibrium potential energy. When the model is transferred to downstream tasks, this initial state becomes unstable due to changes in the potential energy landscape, resulting in an unreliable observation (representation). Therefore, it is inappropriate to predict the model transferability solely based on the current observations (static representations).

%In physics, potential energy is the energy held by an object because of its position relative to other objects, stresses within itself, its electric charge, or other factors. 
% Throughout this paper, the term ``energy'' represents the energy held by the interaction force of different class cluster to push them apart.
\begin{figure}[h]
\begin{center}
\includegraphics[width=1.0\linewidth]{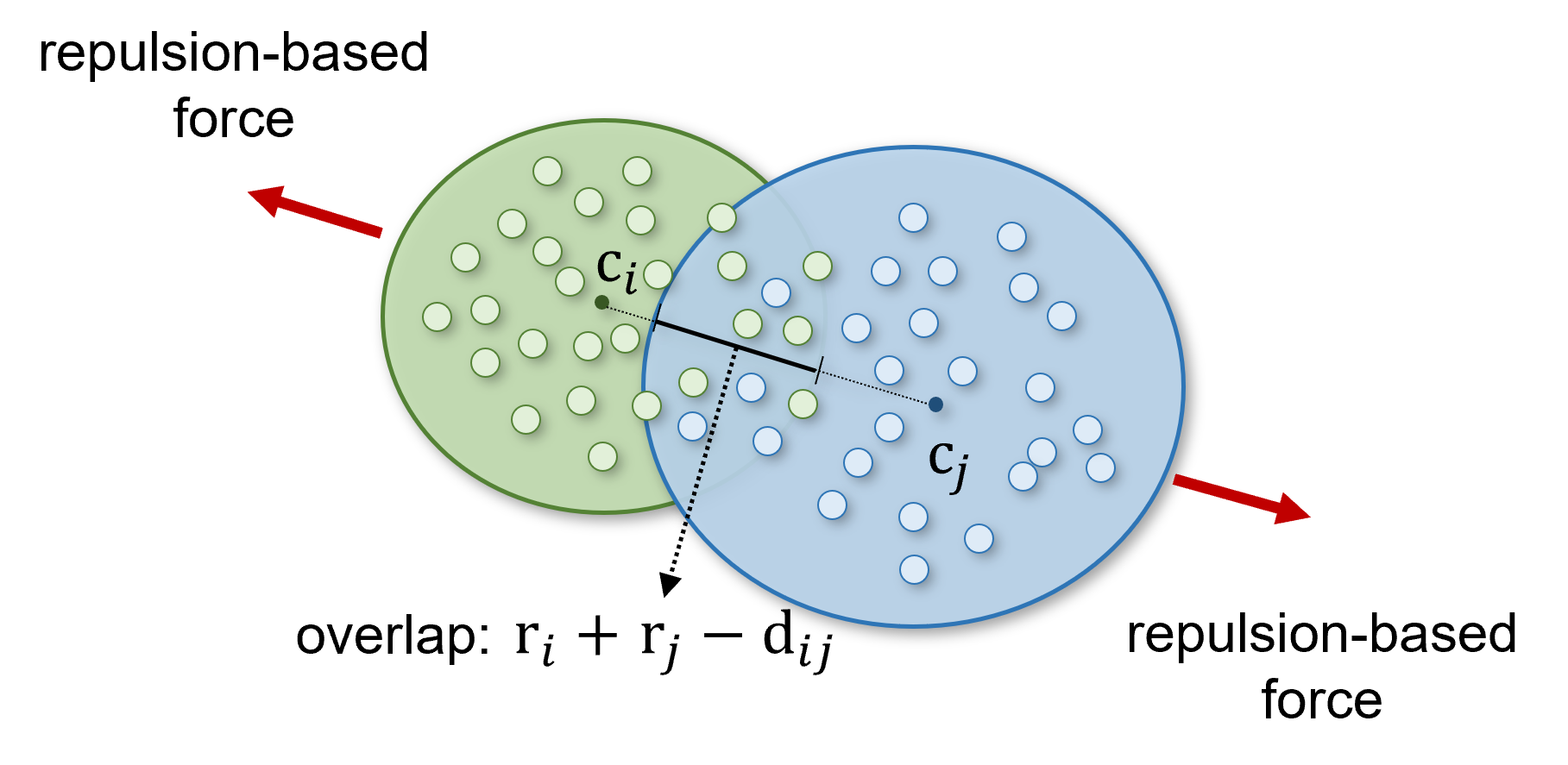}
%\framebox[4.0in]{$\;$}
%\fbox{\rule[-.5cm]{0cm}{4cm} \rule[-.5cm]{4cm}{0cm}}
\end{center}
\caption{By treating the problem as minimizing the potential energy and viewing each cluster as a ball with overlap ($\text{r}_i+\text{r}_j-\text{d}_{ij}$) to another, we can simulate the system's dynamics by releasing the start point and observing how each ball is pushed away under the resulting forces.}
\label{fig:gaussian}
\end{figure}

% In the fine-tuning process, the objective is to push different classes away from each other in the latent space. This can be viewed as creating a force that separates the clusters, which is similar to how objects interact in physics. In this context, the loss function $\mathcal{L}$ defines a potential energy plane $U$ that corresponds to the relative positions of training samples $Z$. Since the current state is unstable and tends to decrease the potential energy \cite{energyprinciple}, it is crucial to capture the movement of the system. To model the changing tendency of the current observation, we need to propose a physical scene for the representation dynamics. Inspired by the concept of elastic potential energy, which arises when an object is deformed under tension or stress, we use a repulsive force to simulate the interactions between different classes, allowing for a more effective and accurate representation of the learning dynamics.
%that is deformed under tension or stress, resulting from a force that repels the object.

During fine-tuning, the objective is to separate different classes in the latent space, which can be seen as a force separating clusters, similar to objects interacting in physics. In this context, the loss function $\mathcal{L}$ creates a potential energy plane $U$ based on the relative positions of training sample features $Z$, with the current state being unstable and favoring a decrease in potential energy \cite{energyprinciple}. Therefore, it is crucial to capture the movement of the system to model the changing tendency of the current observation. To achieve this, we propose a physical scene for representation dynamics based on the concept of elastic potential energy that arises when an object is deformed under tension or stress. We use a repulsive force to simulate the interactions between different classes, improving the effectiveness and accuracy of the learning dynamics.

\subsection{Modeling Representation Dynamics with Mechanical Motion}

% Specifically, each class of points in the embedding space is modeled as a ball with a centroid $\text{c}_i$ as mean features, a radius $\text{r}_i$ as $\lambda\Vert \sigma_i \Vert_{2}$, and a unit mass $m_i$.

We propose a physical modeling approach named Potential Energy Decline (PED), that leverages potential energy to develop a mechanical motion process for representation dynamics. Specifically, each class of feature points $Z_i$ in the embedding space is modeled as a multi-variate Gaussian distribution $\mathcal{N}(\text{c}_i,{\sigma_i}^2)$ with mean feature $\text{c}_i$ and variance ${\sigma_i}^2$. Then we simplify it as a ball with $\text{c}_i$ as its centroid, $\lambda\Vert \sigma_i \Vert_2$ as its radius $\text{r}_i$, and a unit mass $m_i$. The force that repels different features is modeled as an elastic deformation force between the balls, similar to \textit{Hooke's law} \cite{hooke} (\text{i.e.,} $\boldsymbol{F}=k\boldsymbol{x}$). Whenever two balls overlap, a force proportional to the deformation $\boldsymbol{x_e}$ is exerted in the direction of the vector $\boldsymbol{n}$ connecting the centers of the two balls (as shown in Fig. \ref{elastic force}):

\begin{equation}
\begin{split}
\label{elastic force}
& \boldsymbol{x_e}={\text{max}(\text{r}_i +\text{r}_j- \text{d}_{ij}, 0)}, \\
& {\boldsymbol{F}}_{ij}=k\boldsymbol{x_e}\cdot{\boldsymbol{n}},
\end{split}
\end{equation}

where $k$ is a hyper-parameter of elasticity resisting coefficient, and $\text{d}_{ij}$ denotes the distance $\Vert \text{c}_i-\text{c}_j \Vert_{2}$ between centers of the two balls.

As shown in Eq. (\ref{elastic force}), the force between two clusters becomes larger as the overlap becomes larger, and becomes zero when they move apart and no longer overlap. We further model the force from each ball to every other ball and sum up the forces to obtain the joint force ${\boldsymbol{F}}_{i}$ acting on ball $i$. To model the moving tendency, it is revealed by the acceleration $\boldsymbol{a}_i$ by \textit{Newton's} second law of motion \cite{newton}:
\begin{equation}
\begin{split}
\label{motion}
&{\boldsymbol{a}_i}=\frac{{\boldsymbol{F}}_{i}}{m_i}=\frac{\sum\nolimits_{j\neq i}{\boldsymbol{F}}_{ij}}{m_i}.
\end{split}
\end{equation}

In the field of physics, it is often assumed that force remains constant over a very short period of time. Following such a philosophy, we propose a method for simulating the phase position or relative position changes of a system by releasing it within a brief time interval $\Delta t$. The motion equation is then used to compute the position changes.
\begin{equation}
   \overline{Z_i} = Z_i + \Delta Z_i = Z_i+\frac{1}{2}\boldsymbol{a}_i\cdot {\Delta t}^2.
\end{equation}

By applying force to the samples in the system, they are effectively driven towards the direction of decreasing potential energy. By repeating this process multiple times, we obtain an even better system state $\overline{Z}$ with lower potential energy.
% By modeling the initial state as a relative unstable condition and with the tendency to move, our approach can refine the observation of the system and improves the accuracy of predicting model transferability. This physical modeling approach allows us to better understand the dynamics of the system and optimize the transfer learning process.
% \subsection{A Symmetry View of Physical Modeling}
\paragraph{Discuss the feasibility of physical modeling.}
We can view our physics-inspired approach back to the other side of the coin, \textit{i.e.}, conventional gradient-based perspective. We model the elastic potential of the system following \textit{Hooke's law} \cite{hooke} (\text{i.e.,} $U=\frac{1}{2}k\boldsymbol{x}^2$) and the formulation is as follows,
 %By modeling the fine-tuning process as decreasing elastic energy, we formulate the potential energy function of the system as follow,
\begin{equation}
\label{eq:potential energy}
    \begin{split}
    U(Z)&=\sum\nolimits_{i}\sum\nolimits_{j\neq i} \frac{1}{2}k\boldsymbol{x}_{ij}^2\\
    &=\sum\nolimits_{i}\sum\nolimits_{j\neq i} \frac{1}{2}k\text{max}(r_i+r_j-d_{ij}, 0)^2,
    \end{split}
\end{equation}
where $x_{ij}$ describes the overlap of the feature clusters between $Z_i$ and $Z_j$. It is found that the form in Eq. (\ref{eq:potential energy}) is analogous to gradient-based optimization methods that aim to minimize the overlaps of different clusters, which can be viewed as a pairwise loss to enhance class prototype separation in metric learning. 
%The potential energy function can be interpreted as a loss function that minimizes the overlap of different class clusters. This can be viewed as a pairwise metric learning to enhance cluster separation. 
% However, the optimization-based method directly is computationally expensive and not straightforward since it involves updating the entire network and re-extracting the features. In contrast, our approach utilizes a physical modeling approach to decrease the energy potential, which can be incorporated into existing methods to model the tendency of transfer learning dynamics. 
In contrast, our approach offers a more efficient alternative to the optimization-based method by using a physical modeling approach to decrease energy potential, which can be easily integrated into existing methods for transfer learning dynamics.

\subsection{Overall}
Our physical modeling approach provides a refined observation $\overline{Z}$ of the system to take over the initial observation without performing updating the network. The dynamic representation is achieved by mechanical motion and more details of the proposed physics-driven approach can be found in Alg. \ref{alg:algorithm1}.
An arbitrary model selection metric, such as LogMe\cite{logme},
% Then various model selection metrics 
$\mathcal{M}(\overline{Z}, Y)$ can be adopted subsequently to rank the models, \textit{i.e.}, obtaining $\{P_i\}_{i=1}^{N}$. This approach allows us to gain a better understanding of the system's dynamics and boost existing model ranking algorithms.
\begin{algorithm}
\footnotesize
\caption{Algorithm of the proposed Potential Energy Decline (PED)}
\label{alg:algorithm1}
\KwIn{
Model zoo $\{\Phi_i\}_{i=1}^{N}$; Downstream labeled dataset $\mathcal{T}=\{X,Y\}$ including $C$ classes; The hyper-parameter $\lambda$, $\Delta_t$, $k$; Maximum iteration steps $M$, early termination condition $\epsilon$; The model selection metric $\mathcal{M}$;
}
\KwOut{The transferability score $P_i$ for each model in the model zoo.}
\For{$\Phi_i$ in Model zoo}{
\,\, Encode images $X$ to feature embeddings $Z=\Phi_i(X)$ and normalize features with mean $\hat{\mu}$ and standard deviation $\hat{\sigma}$ of ImageNet features:
$Z \leftarrow (Z-\hat{\mu})/\hat{\sigma}$; \\
\While{step $\leq$ M}{
Compute the mean feature as ball center $\text{c}_j$ and standard deviation ${\sigma}_j$ of each class cluster $Z_j$\\
$\text{c}_j=\mathrm{E}[Z_j], \text{r}_j=\lambda||\sigma_j||_2$; \\

Compute distances of the feature clusters $\text{d}_{jl}=||\text{c}_j-\text{c}_l||_2$, $j\neq l\in\{1,...,C\}$; \\
Compute the force and the acceleration of each cluster
$F_{j}=\sum\limits_{l\neq j} k(r_j + r_l - d_{jl})\cdot\frac{\hat{\mu}_j-\hat{\mu}_l}{||\hat{\mu}_j-\hat{\mu}_l||_2}, a_j = \frac{F_j}{m_j}$;\\ 

Simulate the moving process and obtain a more stable state of features $Z$
$z \leftarrow z+\frac{1}{2}a\cdot \Delta_{t}^2, z\in Z_{j}, j\in \{1,...,C\};$  \\
Calculate the ternminal condition
$\omega[step] \leftarrow ||\sum\nolimits_{j=1}^{C}\frac{1}{2} a \cdot \Delta_{t}^2||_1;$ \\
\If{$\omega[step]\leq \epsilon \cdot \omega[0]$}{break;}
step $\leftarrow$ step+1;
}
Features revert back to the original space $Z\leftarrow Z\cdot \hat{\sigma} + \hat{\mu}$;
Feed $Z$ and $Y$ into transferability predicting metric $\mathcal{M}(Z,Y)$ to obtain a score $P\_i$ ;
}
Rank models in ${\Phi\_i}\_{i=1}^{N}$ according to their scores $\{P_i\}_{i=1}^{N}$.;
\end{algorithm}

\begin{table*}[h]
\caption{The experiment results of different transferability metrics on various self-supervised learning models, with the weighted Kendall's $\tau_w$ employed as the ranking correlation protocol. A larger $\tau_w$ represents a better prediction rank order to the ground truth rank. The best results are denoted in bold.}
\centering
\label{tab:table}
\resizebox{\textwidth}{!}
{%
\begin{tabular}{llcccccccccc}
\toprule[1pt]
%\hline
Method & Reference  & Aircraft & Caltech101 & Cars & Cifar10 & Cifar100 & Flowers & VOC & Pets & Food  & DTD\\ \hline
$\mathcal{N}$LEEP \cite{nleep} & CVPR'21& -0.029 & 0.525 & 0.486 & -0.044 & 0.276 & 0.534 & -0.101 & 0.792 & 0574 & 0.641 \\
PARC \cite{parc} & NIPS'21& -0.03 & 0.196 & 0.424 & 0.147 & -0.136 &0.622 & 0.618 & 0.496 & 0.359 & 0.447 \\
% \hline
LogME \cite{logme} & ICML'21 & 0.223 & 0.051 & 0.375 & 0.295 & -0.008 & 0.604 & 0.158 & 0.684 & 0.570  & 0.627 \\
\rowcolor[HTML]{D7F6FF}
LogME+Ours & this paper & \textbf{0.509} & 0.505& 0.516 & 0.511 & \textbf{0.667} & 0.715& \textbf{0.620 } & \textbf{0.795} & 0.650  & 0.780\\
% \hline
% \hline
SFDA \cite{sfda} & ECCV'22 & 0.254 & 0.523  & 0.515 & 0.619 & 0.548 & 0.773  & 0.568 & 0.586 & \textbf{0.685} & 0.749 \\ 
\rowcolor[HTML]{D7F6FF}
SFDA+Ours& this paper & 0.464 & \textbf{0.614}  & \textbf{0.647 } & \textbf{0.673 } & 0.568 &\textbf{0.777}  & 0.583 & 0.462 & 0.581 & \textbf{0.907} \\
GBC \cite{gbc}&CVPR'22 & 0.048 & -0.18 & 0.424 & 0.008 & -0.249 & 0.532 & -0.041 & 0.655 & 0.268 & 0.05\\
\rowcolor[HTML]{D7F6FF}
GBC+Ours & this paper &0.462  & 0.285  & 0.547  & 0.017 & 0.359 & 0.768   & -0.035 & 0.684& 0.402 &0.576 \\
\bottomrule[1pt]
\end{tabular}%
}
\end{table*}

\section{Experiment}

In recent years, self-supervised learning has emerged as a dominant approach in vision pre-training, showing superior transferability compared to supervised learning methods. However, the potential learning dynamics can significantly impact the performance of traditional model transferability prediction metrics for self-supervised pre-trained models. Therefore, in this paper, we analyze the performance on self-supervised learning models to evaluate our proposed approach.

% In the field of vision pre-training, self-supervised learning has gained prominence as a dominant approach in recent years, demonstrating superior transferability compared to supervised learning methods. Nevertheless, the potential learning dynamics can significantly impact the performance of traditional model transferability prediction metrics for self-supervised learning models. In contrast, supervised models exhibit relatively small and less prominent dynamic movement due to the high coincidence degree of the supervised classification objective with downstream tasks. Therefore, in this paper, we analyze the transferability prediction of self-supervised learning methods to evaluate our proposed approach.
% In contrast, due to the supervised classification objective has a high coincidence degree to downstream task, the system movement of supervised model is relative small and less prominent. 

\paragraph{Downstream Dataset.} In this study, we utilize a variety of widely-used datasets for transfer learning in downstream classification tasks, including FGVC Aircraft \cite{aircraft}, Caltech-101 \cite{caltech101}, Standford Cars \cite{cars}, Cifar-10 \cite{cifar}, Cifar-100 \cite{cifar}, DTD \cite{dtd}, Oxford102 Flowers \cite{flowers}, Food-101 \cite{food}, and Oxford-IIIT Pets \cite{pets}. These datasets include diverse and comprehensive characteristics, such as street view, texture, and coarse/fine-grained scenes are suitable for our setting with diversity. 

\paragraph{Pre-trained Model Zoo.} To assess the generality of our method for self-supervised learning models, we consider 12 different types of pre-trained models with ResNet-50 \cite{resnet}, which have been developed using state-of-the-art self-supervised learning methods, including BYOL \cite{byol}, Infomin \cite{infomin}, PCLv1 \cite{pcl}, PCLv2 \cite{pcl}, Selav2 \cite{sela}, InsDis \cite{indis}, SimCLRv1 \cite{simclr-v1}, SimCLRv2 \cite{simclr-v2}, MoCov1 \cite{moco-v1}, MoCov2 \cite{moco-v2}, DeepClusterv2 \cite{deepcluster}, and SWAV \cite{swav} \cite{grill2020bootstrap}. Due to the limited space, some detailed information are provided in Appendix.

\paragraph{Ground Truth Model Rank.} The construction of the ground truth rank $\{G_i\}_{i=1}^{N}$ for model zoo follows the implementation in \cite{sfda,logme}, where a grid search strategy is employed to compute the ground truth performance of each model in downstream task. Specifically, the grid search strategy includes a range of learning rates from the set $\{10^{-1},10^{-2},10^{-3},10^{-4}\}$ and weight decay values from the set $\{10^{-6},10^{-5},10^{-4},10^{-3}\}$. To ensure robustness, each experiment was run with an average of 5 seeds. Obviously, it is observed from the above process that selecting the most suitable model through fine-tuning incurs a significant computational cost in terms of time and GPU resources.

\paragraph{Results of Existing Methods.} We show the experiment results in the Table \ref{tab:table} and it is evident that the state-of-the-art methods encounter significant challenge in predicting transferability of self-supervised models, and even show inability to provide recommendations for some particular datasets, \textit{e.g.,} the Kendall' weights are less than 0. For example, GBC, which achieves impressive performance in supervised learning scenarios by directly measuring the overlap degree of different clusters, exhibits a significant decline in prediction accuracy when tasked with an unstable initial state. Among the techniques evaluated, $\mathcal{N}$LEEP and SFDA exhibit better performances due to their implicit inclusion of a learning process aimed at adapting to downstream tasks. Nevertheless, the limited initial observations still hinder their performance, \textit{e.g.,} the relative low performance in Aircraft. The above experiment results show that predicting the model performances of self-supervised models are not reliable with solely the initial observation and it is of great significance to take the representation dynamics into consideration.

\paragraph{Results of Our method.} To evaluate the efficacy of our approach, we integrate our method upon different state-of-the-art transferability metrics, including evidence-based LogME \cite{logme}, discrimination-based SFDA \cite{sfda}, separation-based GBC \cite{gbc}. Through taking the dynamics representations into consideration, the performance combined with ours approach show obvious gains in many downstream scenes. For instance, our approach yields a significant performance gain of +0.675 and +0.608 compared to LogME and GBC on Cifar100, respectively. Even though SFDA has specifically designed Confix to alleviate the dynamics of fine-tuning by augmenting hard examples, it remains orthogonal to our method. For example, it achieves a gain of +0.210 and +0.158 on Aircraft and DTD, respectively. Although existing methods have achieved remarkable performance in experimental results (B), achieving above 0.6 in Kendall weight, our method can still provide diverse benefits upon different metrics. Our experiments confirm the effectiveness of our physics-inspired modeling approach and highlight the significance of considering representations in self-supervised models.

\section{Ablation Study}
In this section, we perform an ablation study of our method on downstream datasets of Cars, Flowers, and DTD. Specifically, we investigate the impact of the hyper-parameters and implementation details of our method. Through these experiments, we aim to gain a deeper understanding of our method's performance and identify key factors that contribute to its effectiveness.

\subsection{Period of Time}
\begin{figure}[h]
\begin{center}
\includegraphics[width=1.0\linewidth]{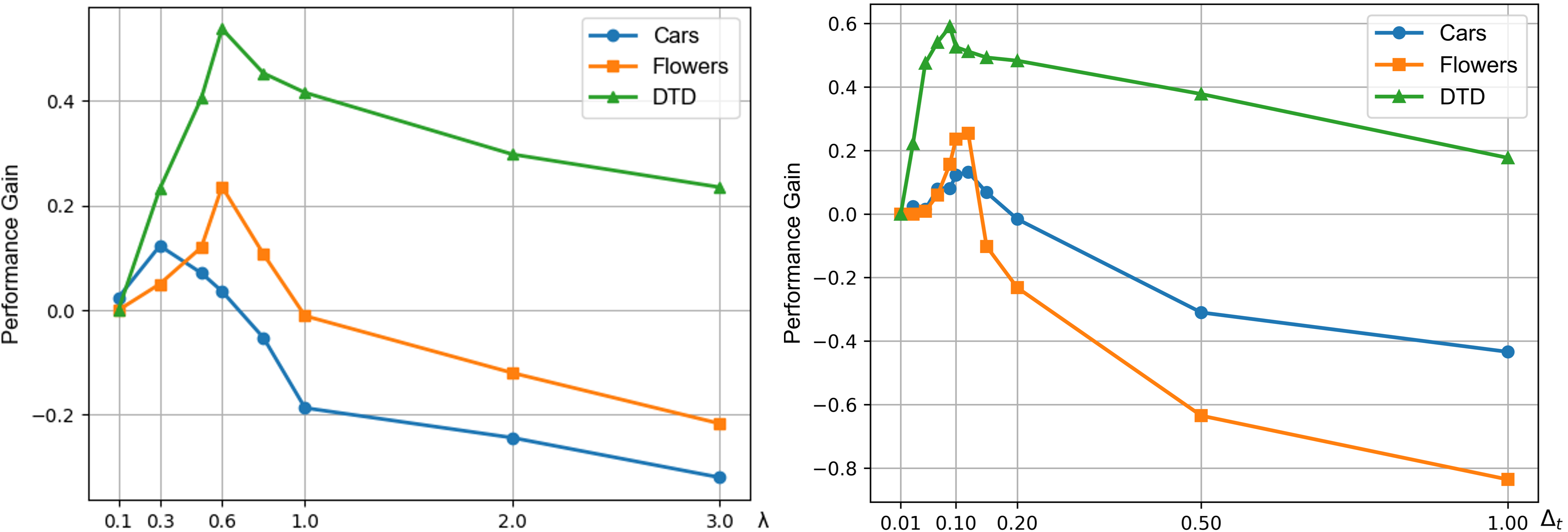}
%\framebox[4.0in]{$\;$}
%\fbox{\rule[-.5cm]{0cm}{4cm} \rule[-.5cm]{4cm}{0cm}}
\end{center}
\caption{\label{fig:lambda}The performance gain in Kendell $\tau$ with respect to different $\lambda$ and $\Delta_t$.}
\end{figure}

The period of time $\Delta_t$ is a crucial factor that determines the degree of the dynamic process. In Fig. \ref{fig:lambda}, we conduct ablation experiments to investigate the effect of varying $\Delta_t$.  
The results demonstrate that increasing the time of movement drives the clusters to decline the potential energy, which in turn leads to improved observation and performance gain.  

In physics, force conditions are typically approximated as constant over a short period of time. However, as the time period increases, the clusters move further apart, causing the force to change significantly and introducing errors into the physical assumption. Consequently, we set the time period to a small value of $0.1$ for all experiments. 
% Our findings indicate that controlling the dynamic process can enhance the model's performance and that selecting an appropriate time period is crucial in achieving optimal results.

\subsection{The Radius Coefficient $\lambda$}

As we represent the features of each cluster as a Gaussian distribution, and simplify it as a ball in a physical view, we adopt the radius coefficient $\lambda$ controls the degree of modeling the overlaps among different clusters and set hyper-parameter $k$ to be 1.0 by default.

Through ablation experiments presented in Fig. \ref{fig:lambda}, we interestingly observed that setting $\lambda$ to $0.3$ yields in better performance in datasets with significant data cluster overlaps, such as Cars, whereas setting $\lambda$ to 0.6 was more effective for datasets with less cluster differences, such as Cifar10. Consequently, we use the set of $\{0.3,0.6\}$ as candidate radius coefficients for all downstream tasks. Our findings indicate that controlling the dynamic process can enhance the model's performance and that selecting an appropriate radius coefficient is crucial in achieving optimal results.

\subsection{Multiple Step Moving}
\begin{figure}[h]
\begin{center}
\includegraphics[width=1.0\linewidth]{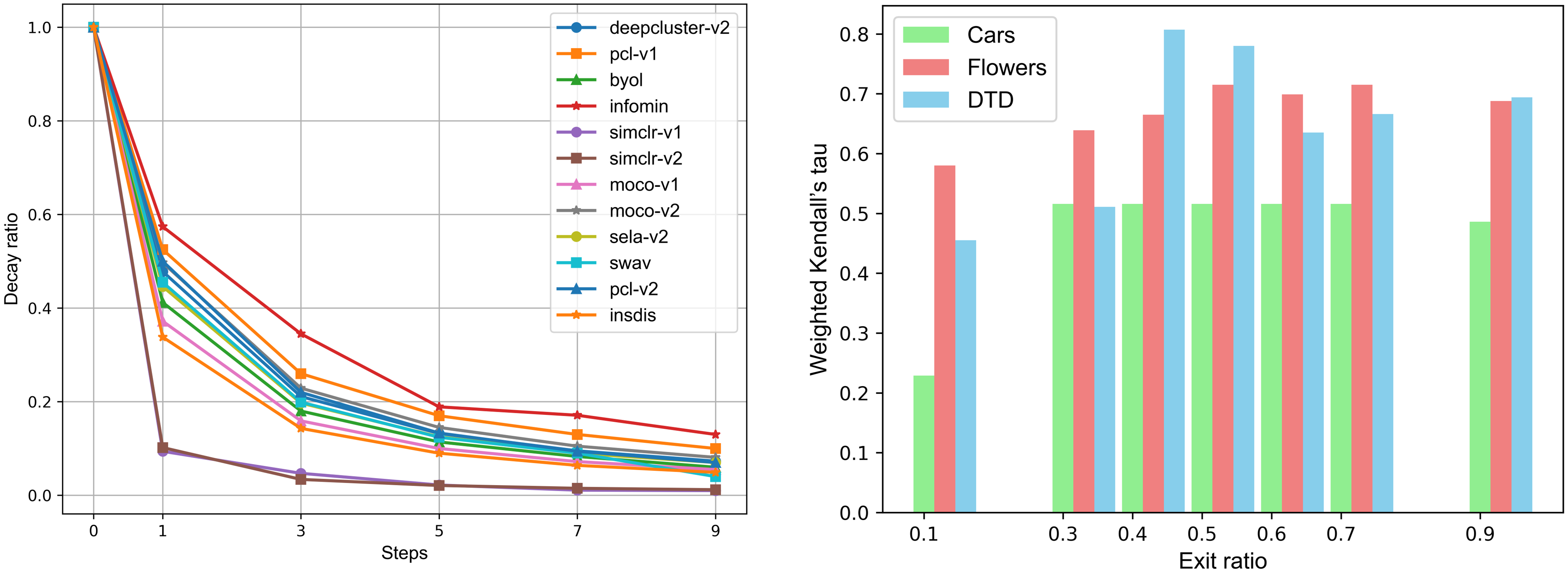}
%\framebox[4.0in]{$\;$}
%\fbox{\rule[-.5cm]{0cm}{4cm} \rule[-.5cm]{4cm}{0cm}}
\end{center}
\caption{\label{fig:iteration} The left picture is the decay curve of force changing with steps $M$ and the right picture shows the influence of different end condition to the performance.}
\end{figure}

To decrease the potential energy and achieve the desired effect, we employ a multi-step moving process, where the exit ratio $\epsilon$ and the maximum number of steps $M$ determine the end condition. As shown in Fig. \ref{fig:iteration}, it is evident that the force decreases rapidly within a few steps, indicating that the clusters are being pushed apart and the interaction force is decreasing. 

According to the results, we set the attenuation condition and the maximum number of steps to be 5 to terminate the moving process when the current force is less than $\epsilon=0.5$ of the initial force in the experiments. The multiple step moving approach allows us to model the movement in a short period of time, while taking into account the updated feature positions and re-calculating the interaction force.

\section{Visualization and Analysis}
In this section, we present visualization results to assess the effectiveness of our proposed method and to gain a deeper understanding of its mechanisms.

\subsection{Different Cluster Change}
\begin{figure}[h]
\begin{center}
\includegraphics[width=1.0\linewidth]{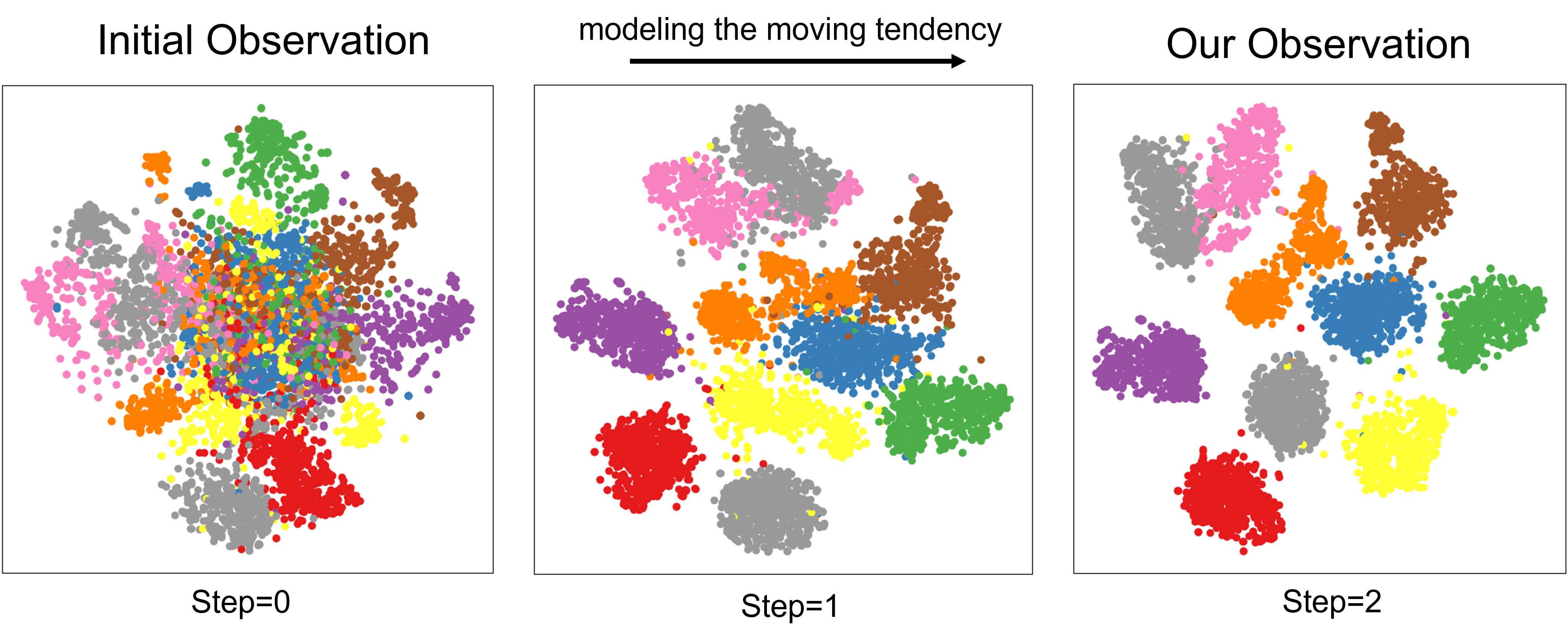}
%\framebox[4.0in]{$\;$}
%\fbox{\rule[-.5cm]{0cm}{4cm} \rule[-.5cm]{4cm}{0cm}}
\end{center}
\caption{\label{tsne}The dynamic representation of our modeling process on self-supervised models.}
\end{figure}

The visualization in Fig. \ref{tsne} reveals that the initial state of clusters, encoded by self-supervised pre-trained models, is not well-distributed. While the different clusters should be apart from the others when transferring to downstream task. Therefore, this highlights the potential for improvement that our method offers. To further investigate the underlying processes, we take the BYOL model on Cifar100 dataset and employ the t-SNE algorithm \cite{tsne} to visualize our dynamic modeling process.

As shown in Fig. \ref{tsne}, the initial observations are rather chaotic and samples are not clearly separated by class in the t-SNE visualization, which is due to the lack of class information in the self-supervised pre-training. Therefore, predicting the transferability of the initial state is unreliable. However, our dynamic modeling process improves the separability of the feature clusters, achieving a similar effect to fine-tuning without requiring network updates. Overall, the visualization results provide strong evidence in support of the effectiveness of our proposed method, and offer insights into the underlying processes that contribute to its success.

\subsection{Model Rank}
\begin{figure}[h]
\begin{center}
\includegraphics[width=1.0\linewidth]{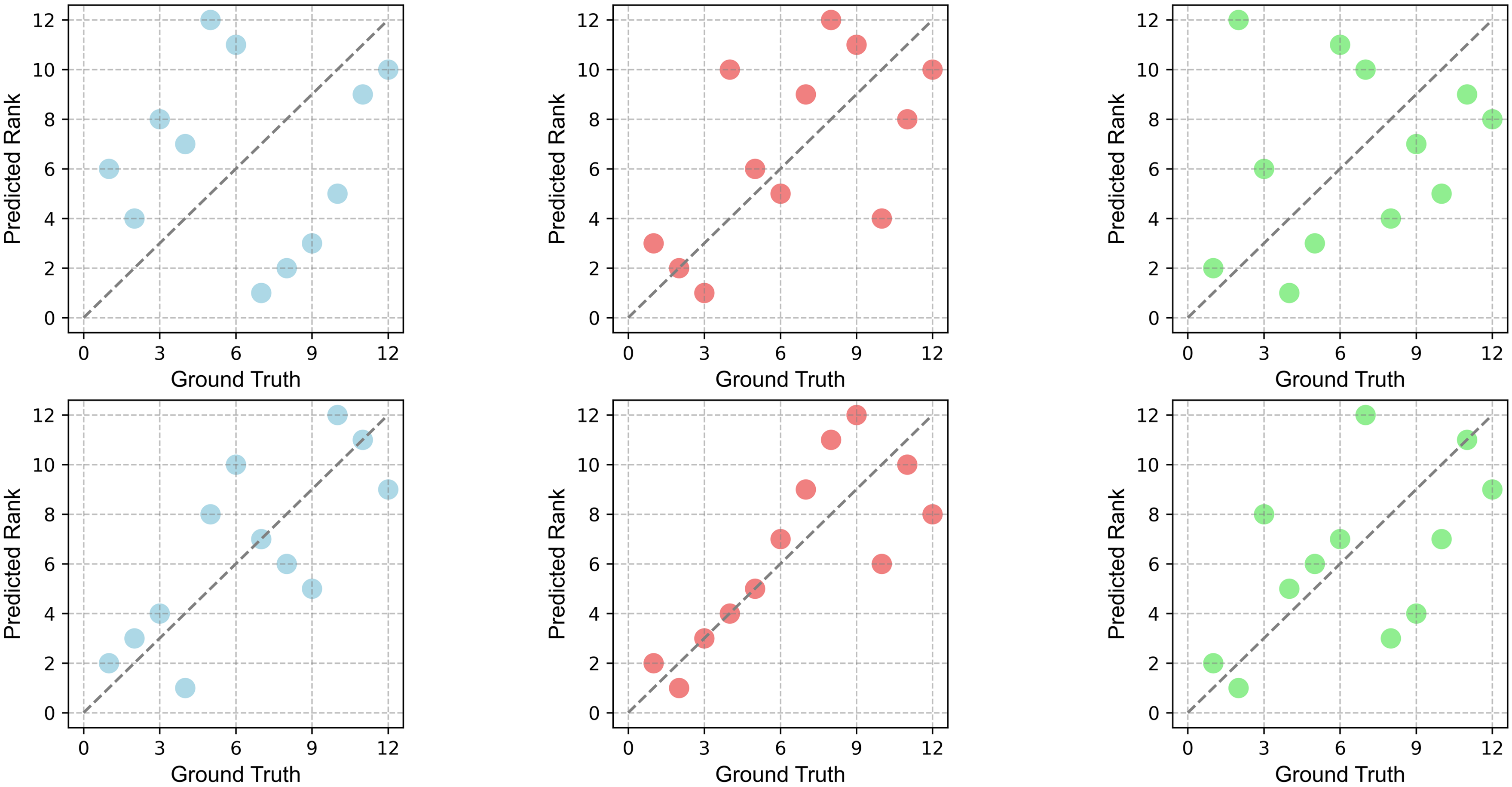}
%\framebox[4.0in]{$\;$}
%\fbox{\rule[-.5cm]{0cm}{4cm} \rule[-.5cm]{4cm}{0cm}}
\end{center}
\caption{\label{fig:model rank}We present a visualization of the variation in model ranking concerning the prediction of transferability scores, utilizing both our own observations and initial observations. The depicted progression spans three datasets: DTD, Flowers, and Cars, arranged from left to right.}
% \vspace{-0.5cm}
\end{figure}

By utilizing our proposed method, we are able to refine models that have a relatively low transferability score as a result of initial unstable observations. To evaluate the efficacy of our approach, we generated a visualization of the model rank comparisons between the initial observation and our refined observation, as shown in Fig. \ref{fig:model rank}.

The results demonstrate a significant improvement in the calibration of model rankings when utilizing our refined observation, where the optimal model is positioned near the reference line in the figure. Notably, some models that begin with an unfavorable starting point can be rapidly improved to achieve a superior ranking, highlighting the effectiveness of our refinement methodology.

% \subsection{Comparing to Optimization method}

% We also include the comparisons with backpropgation-based method for fine-tuning. It could be inferred that predicting the model performance after the converged fine-tuning model will obtain better performance. 

\section{Efficiency Analysis}
In the results of our experiment, we have demonstrated the effectiveness of our proposed method in enhancing various transferability prediction techniques. Additionally, we highlight in this section that our approach exhibits computational efficiency in terms of algorithmic complexity and practical running time.
\begin{table}[h]
\begin{center}
\caption{The comparisons of running time on Flowers.}
\label{table:running time}
\resizebox{\linewidth}{!}{
\begin{tabular}{c|ccccc|c}
\toprule[1pt]
Metrics & PARC & LogME & NLEEP & SFDA  & GBC  & Ours \\ \hline
Running Time  & 27s  & 8s & 392s  & 82s  &  11s   & 2s \\
\toprule[1pt]
\end{tabular}}
\end{center}
\vspace{-0.5cm}
\end{table}
%\paragraph{Complexity.} 
Our method is computationally efficient due to the simplification in physics that we have adopted. Specifically, we model the same class features as a whole ball and consider the stress of the class center exclusively, resulting in a computational complexity of $\mathcal{O}(C^2D)$, where $C$ represents the number of classes and $D$ denotes the feature dimension. Consequently, our method produces a relatively small time overhead in comparison to the transferability prediction process. We present our experimental findings on the running time of our approach in Table \ref{table:running time}. Notably, our method contributes minimal overhead to the transferability prediction process, further attesting to its effectiveness and efficiency in computation.

\section{Conclusions and Future Work}
This paper presents a fresh insight to reframe transfer learning as a process of decreasing the system potential energy. To this end, we propose physically motivated modeling technique that effectively captures the dynamics of representations. Despite being a simplified physical modeling approach, our method consistently boosts the existing metrics for ranking the self-supervised pre-trained models. In the future work, we intend to enhance the sophistication of our physical model by incorporating adaptive hyper-parameters and expanding its applicability to more transfer learning scenes. We hope our work will shed light on the representation dynamics in transfer learning and inspire further research in this field.

\section*{Acknowledgement}
\label{acknowlegement}
This work was supported by the National Natural Science Foundation of China under Grant 62088102, and in part by the PKU-NTU Joint Research Institute (JRI) sponsored by a donation from the Ng Teng Fong Charitable Foundation.

{\small
\bibliographystyle{ieee_fullname}
\bibliography{egpaper_for_review}
}

\clearpage
\appendix

\section{Compared Methods}
In this paper, we evaluate the efficacy of our proposed method by applying it to three distinct transferability prediction metrics, $\textit{i.e.,}$ LogME \cite{logme}, GBC \cite{gbc}, and SFDA \cite{sfda}. To enhance understanding of their underlying principles and mechanisms, we provide detailed descriptions of these metrics in this section.

\paragraph{LogME \cite{logme}.} LogME is an evidence-based metric, which uses the marginal evidence to measure the transferability. Unlike the approach in \cite{h-score}, LogME does not directly minimize the Gaussian-based log-likelihood. Instead, it adopts Bayesian averaging to address the overfitting problem:
\begin{equation*}
    p(y|F)=\int p(w)p(y|F,w)\text{d}w,
\end{equation*}
% where $p(w)$ is defined as a Gaussian prior and $p(y|F,w)$ depicts a Gaussian likelihood. 
where $p(w)$ and $p(y|F,w)$ are modeled as two Gaussian distributions specified by two positive parameters.
$p(y|F)$ denotes the probability density of the compatibility between features $F$ and labels $y$, which is based on the marginal evidence of the target task.

\paragraph{SFDA \cite{sfda}.} SFDA is a class-discrimination based metric, which utilizes a Fisher Discriminant Analysis (FDA) approach and propose ConfMix to produce hard-negative samples in a self-challenge manner. The aim of SFDA is to find a transformation $U$ to maximize between scatter of classes and minimize within scatter of each class:
\begin{equation*}
    U=\text{arg }\mathop{\text{max}}_{U}=\frac{|U^\top S_B U|}{|U^\top(1-\lambda) S_W+\lambda I)U|},
\end{equation*}
where $S\_B$ and $S\_W$ are the between and within class scatter matrix. The solution can be solved with a close-form solution and then SFDA acquires transformed feature $\{\hat{x}\_n=U^{\top}x_n\}\_\{n=1\}^N$. Finally, SFDA adopts Bayes theorem to obtain the score function $\delta_c(\hat{x}_n)$ and use the probability likelihood to measure the transferability score.

\begin{equation*}
    \delta_c(\hat{x}_n)=\hat{x_n}UU\top\mu_c-\frac{1}{2}\mu_cUU\top\mu_c+\text{log}q_c.
\end{equation*}
% Then the transferability score can be obtained

\paragraph{GBC \cite{gbc}.} GBC is a class-separation based metric that employs the Gaussian Bhattacharyya Coefficient (GBC) to estimate the pairwise class separability.

\begin{equation*}
\begin{split}
    &\text{GBC}=-\sum_{i\neq j}\text{exp}(-\text{BC}(i,j))\\
    &\text{BC}(i,j)=\frac{1}{8}(\mu_{c_i}-\mu_{c_j})^\top\Sigma^\top(\mu_{c_i}-\mu_{c_j}) \\
    &\quad\quad\quad\quad+\frac{1}{2}\text{ln}(\frac{|\Sigma |}{\sqrt{|\Sigma_{c_i} | |\Sigma_{c_j} |}}),
\end{split}
\end{equation*}
where $\mu$ and $\Sigma$ represent the distribution mean and variance of the corresponding class, and coefficient $\text{BC}(i,j)$ denotes the overlaps between classes $i$ and $j$. The final transferability score is based on the overlaps of all classes by summing up the pairwise negative exponential coefficients.
% \begin{equation}
%     \text{GBC}=-\sum_{i\neq j}\text{BC}(c)
% \end{equation}
\section{Implementation Details}
The implementation details are presented in the section of experiment setup and ablation study. Additionally, we present supplementary studies in this section.

\begin{table*}[t]
\small
\caption{The supplementary experiment results of different transferability metrics on various self-supervised learning models under grid-search of hyper-parameter $k$, showing that our method still has further potential with fine-grained tuning on hyper-parameter.}
\centering
\label{tab:k}
%\resizebox{\textwidth}{!}
{%
\begin{tabular}{llcccccccccc}
\toprule[1pt]
%\hline
Self-Supervised & Reference  & Aircraft & Caltech101 & Cars & Cifar10 & Cifar100 & Flowers & VOC & Pets & Food & DTD \\ \hline
$\mathcal{N}$LEEP \cite{nleep} & CVPR'21& -0.029 & 0.525 & 0.486 & -0.044 & 0.276 & 0.534 & -0.101 & 0.792 & 0574 & 0.641 \\
PARC \cite{parc} & NIPS'21& -0.03 & 0.196 & 0.424 & 0.147 & -0.136 & 0.622 & 0.618 & 0.496 & 0.359 & 0.447\\
% \hline
LogME \cite{logme} & ICML'21 & 0.223 & 0.051 & 0.375 & 0.295 & -0.008 & 0.604 & 0.158 & 0.684 & 0.570  & 0.627\\
\rowcolor[HTML]{D7F6FF}
LogME+Ours & this paper & 0.509 & 0.611 & 0.624 & 0.633  & 0.668 & 0.728 & \textbf{0.781} & \textbf{0.795} & \textbf{0.737}  & 0.837\\
% \hline
% \hline
SFDA \cite{sfda} & ECCV'22 & 0.254 & 0.523  & 0.515 & 0.619 & 0.548 & 0.773  & 0.568 & 0.586 & 0.685 & 0.749\\ 
\rowcolor[HTML]{D7F6FF}
SFDA+Ours& this paper & 0.505 & \textbf{0.661}  & \textbf{0.666} & \textbf{0.741} & \textbf{0.744} & \textbf{0.798}  & 0.613  & 0.592 & 0.689 & \textbf{0.907}\\
GBC \cite{gbc}&CVPR'22 & 0.048 & -0.18 & 0.424 & 0.008 & -0.249 & 0.532 & -0.041 & 0.655 & 0.268 & 0.05\\
\rowcolor[HTML]{D7F6FF}
GBC+Ours & this paper &\textbf{0.549}  & 0.340  & 0.629 & 0.149 & 0.431  & 0.779  & 0.552 & 0.758 & 0.672 &0.611\\
\bottomrule[1pt]
\end{tabular}%
}
\end{table*}

\begin{figure*}[h]
\begin{center}
\includegraphics[width=1.0\linewidth]{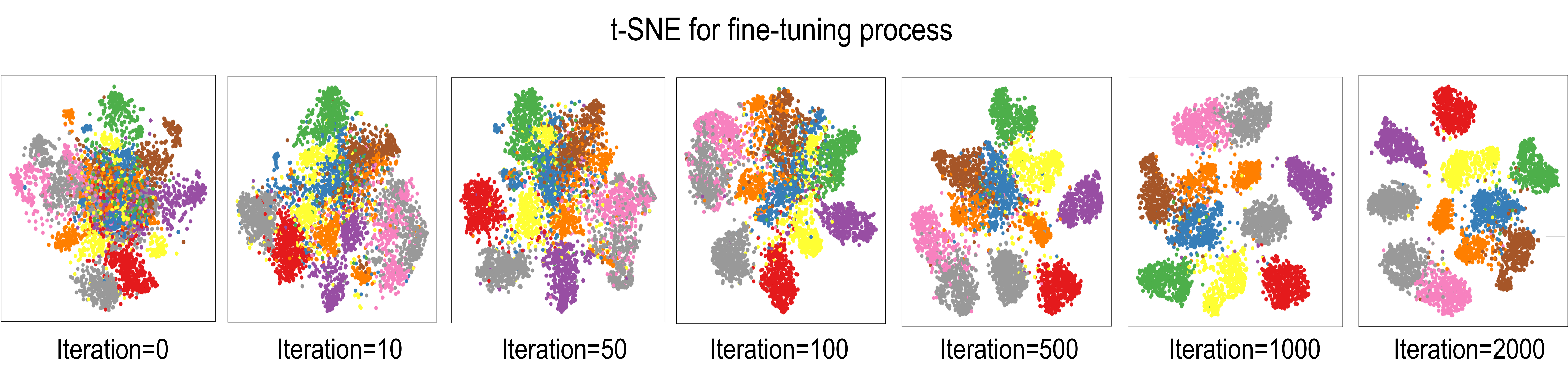}
%\framebox[4.0in]{$\;$}
%\fbox{\rule[-.5cm]{0cm}{4cm} \rule[-.5cm]{4cm}{0cm}}
\end{center}
\caption{\label{fig:fine-tuning} The t-SNE visualization of dynamic feature representation achieved through fine-tuning.}
% \vspace{-0.5cm}
\end{figure*}

\paragraph{Hyper-parameter $k$.} The hyper-parameter $k$ denotes the elastic coefficient of the repulsive-based elastic force and a higher value of $k$ yields a stronger force, as shown in Section 3.3. Since that the elastic hyper-parameter $k$ is coupled with the radius coefficient $\lambda$, we set the default value of the hyper-parameter $k$ to 1.0 and adjust $\lambda$ in our experiments accordingly. As suggested by \cite{sfda}, we further conduct a grid search on $k$ for optimal performance using values of [0.6, 0.8, 1.0, 1.2, 1.5, 2.0] and the results are presented in Table \ref{tab:k}. %The results indicate that the performance can still be further improved with refined hyper-parameters.
%According to \cite{sfda}, hyper-parameter can be searched from a grid search. In Table \ref{tab:k}, we provide the results of a grid search for the elastic hyper-parameter $k$, using values of [0.6, 0.8, 1.0, 1.2, 1.5, 2.0], The results indicate that the performance can still be further improved with refined hyper-parameters.

% In fact, the hyper-parameter $k$ can be adjusted to further enhance the performance. Table \ref{tab:k} displays the results obtained from a grid search ([0.6, 0.8, 1.0, 1.2, 1.5, 2.0]) for the elastic hyper-parameter $k$, implying that our performance can be further improved with refined parameters.

\begin{figure}[h]
\begin{center}
\includegraphics[width=0.8\linewidth]{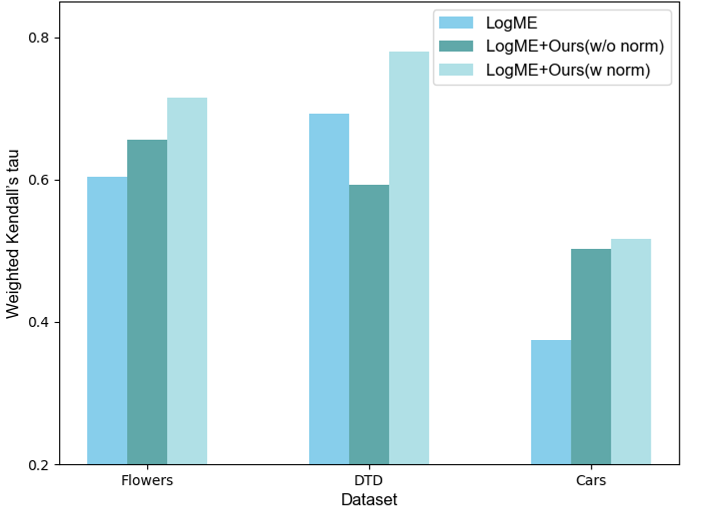}
%\framebox[4.0in]{$\;$}
%\fbox{\rule[-.5cm]{0cm}{4cm} \rule[-.5cm]{4cm}{0cm}}
\end{center}
\caption{\label{fig:norm} The influence of feature pre-processing.}
\vspace{-5pt}
\end{figure}

\paragraph{Feature Pre-processing.}
In our implementation, we adopt a pre-processing step to analyze the motion in the embedding space. Specifically, downstream features are normalized with ImageNet feature mean and standard deviation, based on a subset of 50,000 images. To evaluate the impact of normalization on the modeling process, we present our findings in Fig \ref{fig:norm}.
% \begin{table}[h]
% \begin{center}
% \caption{The additional running time of our method.}
% \label{table:norm}
% \resizebox{\linewidth}{!}{
% \begin{tabular}{c|ccc}
% \toprule[1pt]
% Metrics & LogMe & +Ours(w/o norm) & +Ours (w norm)\\ \hline
% % PACS
% % GTA5 
% Kendell   &     0.1    & 0.3 & 0.6        \\
% %Duke to Market       & 33.7 & 34.5 & 35.4 & 35.6 & 35.4 & 35.4 & 35.6 \
% \toprule[1pt]
% \end{tabular}}
% \end{center}
% \end{table}
% \begin{table}[h]
% \begin{center}
% \caption{The influence of feature pre-processing.}
% \label{table:norm}
% %\resizebox{\linewidth}{!}{
% \begin{tabular}{c|ccc}
% \toprule[1pt]
% Metrics &               Flowers & DTD & Cars \\ \hline
% LogME                 & 0.604    & 0.692 & 0.375          \\
% LogME+Ours (w/o norm) & 0.656    & 0.592 & 0.603 \\
% LogME+Ours (w norm) & 0.715    & 0.780 & 0.516 \\
% %Duke to Market       & 33.7 & 34.5 & 35.4 & 35.6 & 35.4 & 35.4 & 35.6 \
% \toprule[1pt]
% \end{tabular}
% %}
% \end{center}
% \end{table}
Through the normalization, the feature values in each dimension are largely normalized in a certain region (\textit{e.g.,} $[-3\sigma,3\sigma]$ due to the property of Gaussian distribution), creating a suitable condition for physical modeling. We discovered that the normalization can prevent the occurrence of highly imbalanced dimensions caused by the divergence in numerical value and stabilize the physical modeling process.
% It could be seen that the improvement will be narrowed without normalization. 

% In fact, as the physical environment between embedding and real world physical space, we need to normal the feature with different dimensions to avoid too ill-condition unbalances of each dimension in the high-dimension space. Through the normalization, the feature in each dimension are normalized in $[-3\sigma,3\sigma]$ with a large probability due to the property of Gaussian distribution, creating a better space for modeling the dynamics representation in the perspective of physics.

\paragraph{Maximum Phase Position.} We calculate the phase position $s$ of each cluster using the motion equation. In our implementation, we set a maximum phase position constraint, $\text{i.e.,}$ $\text{min}(s, x_e)$. This constraint ensures that the force decreases accordingly as the movement $s$ surpasses the overlap $x_e$ between two clusters. By adding this boundary condition to the motion equation, we can enhances the algorithm's robustness and avoid extreme situations.

% \section{Supervised Pre-trained Models.}
% Despite our method is designed for self-supervised method, we find that we can also contribute to boost the existing methods in predicting supervised pre-trained models. Here we also conduct experiments on four datasets that is challenge to current model selection methods. The experiment results are shown in Table \ref{}.
% \begin{figure*}[h]
% \begin{center}
% \includegraphics[width=0.9\linewidth]{image/fine-tuning2.png}
% %\framebox[4.0in]{$\;$}
% %\fbox{\rule[-.5cm]{0cm}{4cm} \rule[-.5cm]{4cm}{0cm}}
% \end{center}
% \caption{\label{fig:fine-tuning} The dynamic feature representation achieved through fine-tuning.}
% \vspace{-0.5cm}
% \end{figure*}

\begin{table*}[h]
\caption{The ground truth results of the 12 self-supervised pre-trained models on 10 downstream tasks.}
\centering
\label{tab:gt}
%\resizebox{\textwidth}{!}
\begin{tabular}{lcccccccccc}
\toprule[1pt]
\multicolumn{1}{l}{Self-Supervised} & Aircraft & Caltech101 & Cars & Cifar10 & Cifar100 & Flowers & VOC & Pets & Food & DTD \\ \hline
BYOL \cite{byol} & 82.1 & 91.9 & 89.83 & 96.98 & 83.86 & 96.8 & 85.13 & 91.48 & 85.44 & 76.37 \\
Deepclusterv2 \cite{deepcluster} & 82.43 & 91.16 & 90.16 & 97.17 & 84.84 & 97.05 & 85.38 & 90.89 & 87.24 & 77.31 \\
Infomin \cite{infomin} & 83.78 & 80.86 & 86.9 & 96.72 & 70.89 & 95.81 & 81.41 & 90.92 & 78.82 & 73.74 \\
InsDis \cite{indis}& 79.7 & 77.21 & 80.21 & 93.08 & 69.08 & 93.63 & 76.33 & 84.58 & 76.47 & 66.4 \\
MoCov1 \cite{moco-v1}& 81.85 & 79.68 & 82.19 & 84.15 & 71.23 & 94.32 & 77.94 & 85.26 & 77.21 & 67.36 \\
MoCov2 \cite{moco-v2}& 83.7 & 82.76 & 85.55 & 96.48 & 71.27 & 95.12 & 78.32 & 89.06 & 77.15 & 72.56 \\
PCLv1 \cite{pcl}& 82.16 & 88.6 & 87.15 & 86.42 & 79.44 & 95.62 & 91.91 & 88.93 & 77.7 & 73.28 \\
PCLv2 \cite{pcl}& 83.0 & 87.52 & 85.56 & 96.55 & 79.84 & 95.87 & 81.85 & 88.72 & 80.29 & 69.3 \\
Sela-v2 \cite{sela}& 85.42 & 90.53 & 89.85 & 96.85 & 84.36 & 96.22 & 85.52 & 89.61 & 86.37 & 76.03 \\
SimCLRv1 \cite{simclr-v1}& 80.54 & 90.94 & 89.98 & 97.09 & 84.49 & 95.33 & 83.29 & 88.53 & 82.2 & 73.97 \\
SimCLRv2 \cite{simclr-v2}& 81.5 & 88.58 & 88.82 & 96.22 & 78.91 & 95.39 & 83.08 & 89.18 & 82.23 & 94.71 \\
Swav \cite{swav}& 83.04 & 89.49 & 89.81 & 96.81 & 83.78 & 97.11 & 85.06 & 90.59 & 87.22 & 76.68 \\
\bottomrule[1pt]
\end{tabular}%
\vspace*{10in}
\end{table*}

\section{Comparing to Fine-tuning} In this paper, we have shown the effectiveness of our method without the need for fine-tuning, Additionally, we highlight the advantages of our method over fine-tuning in this section. The t-SNE visualization in Fig. \ref{fig:fine-tuning} reveals that during the initial stage of fine-tuning, the clusters are not well separated due to the random initialized classifier layer and further adaptation to downstream tasks is required. For comparison, the t-SNE visualization of our approach is shown in Section 6.1.

\begin{figure}[h]
\begin{center}
\includegraphics[width=0.75\linewidth]{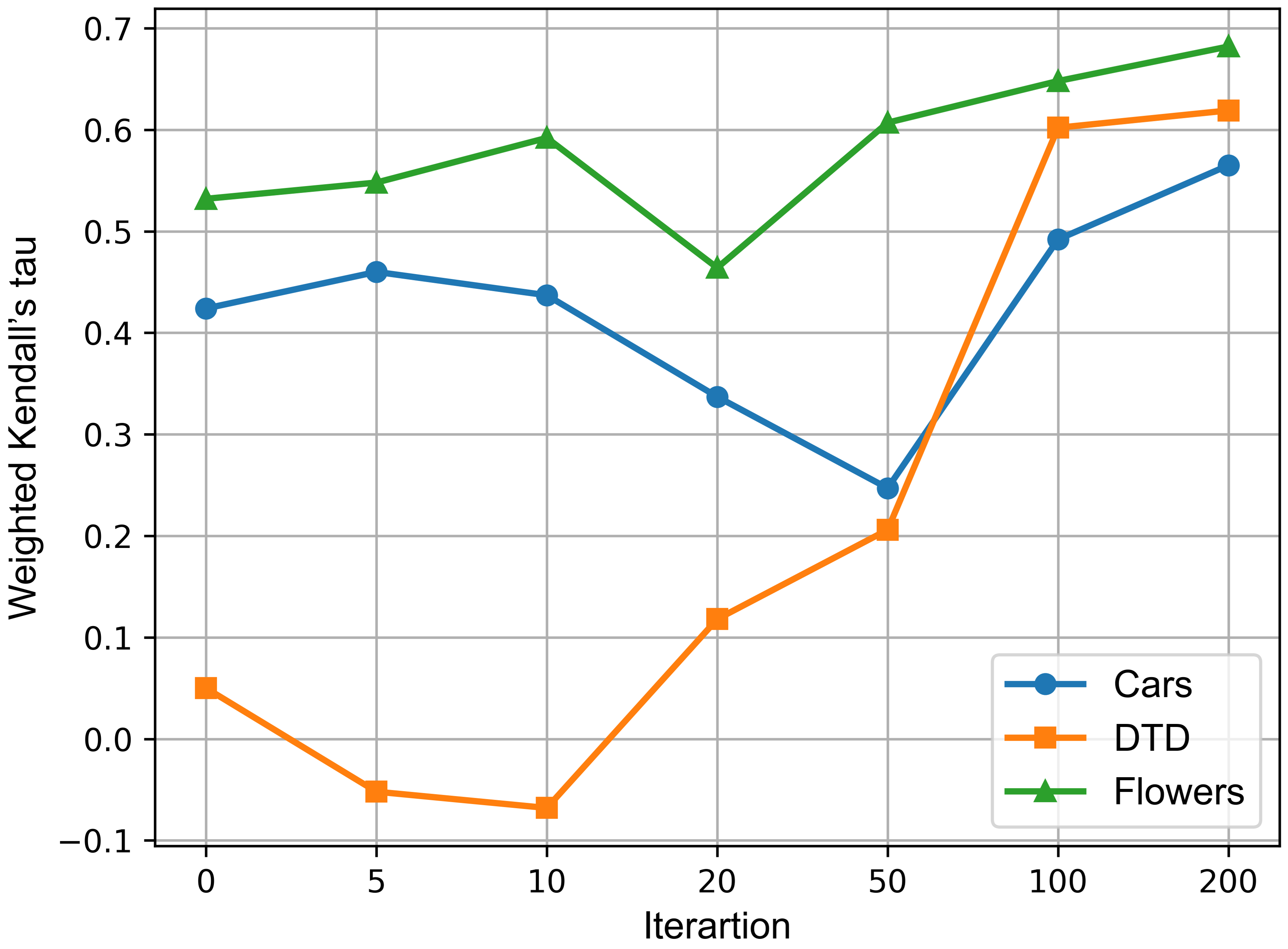}
%\framebox[4.0in]{$\;$}
%\fbox{\rule[-.5cm]{0cm}{4cm} \rule[-.5cm]{4cm}{0cm}}
\end{center}
\caption{\label{fig:ft_dynamic} The ranking performance of different fine-tuning iterations.}
% \vspace{-0.5cm}
\end{figure}

We display the ranking performance achieved through fine-tuning in Fig. \ref{fig:ft_dynamic}, which reveals a performance pattern of initial decline followed by improvement. This suggests that fine-tuning requires multiple iterations to adapt to new tasks for learning the classifier layer. In contrast, our proposed physics-inspired method can simulate the dynamic feature representation without the need for this adaptation process. 

Furthermore, fine-tuning involves a grid search strategy to select the best hyper-parameters, and fine-tuning the entire model on the downstream dataset. This process requires testing 30 hyper-parameter setups, with each training process consisting of 5000 iterations taking 16 minutes, making it more time-consuming compared to our physics-driven approach.

\section{Ground Truth Results}
We obtained the ground truth results by fine-tuning the models using a grid-search strategy, following the the implementation of \cite{sfda,logme}. More information on this process can be found in Section 4. In Table \ref{tab:gt}, we present the ground truth results of the 12 self-supervised learning models and 10 downstream tasks.
% Although the proposed method upon various selection metrics shows effectiveness in measuring transferability of pre-trained models, there are still several future works are worth investigating in pursuit of a more universal metric. Firstly,
% an interesting future work would be how to extend it in other types of target
% tasks such as segmentation tasks with specific modeling. Secondly, we provide a simplified physical modeling and provide a fresh insight, which can be further refined to be more adaptive to different transfer learning scenes. We hope our work will shed light on the fine-tuning dynamics in transfer learning and inspire more work for transfer learning. 

\end{document}